\documentclass[sigconf]{acmart}

\renewcommand\footnotetextcopyrightpermission[1]{} % removes footnote with conference information in first column
\AtBeginDocument{%
  \providecommand\BibTeX{{%
    \normalfont B\kern-0.5em{\scshape i\kern-0.25em b}\kern-0.8em\TeX}}}

\begin{document}

%
% The "title" command has an optional parameter, allowing the author to define a "short title" to be used in page headers.
\title{Kinetic Market Model: An Evolutionary Algorithm}

\author{Evandro Luquini}
\author{Nizam Omar}

%
% The abstract is a short summary of the work to be presented in the article.
\begin{abstract}
This research proposes the econophysics kinetic market model as an evolutionary algorithm's instance. The immediate results from this proposal is a new replacement rule for family competition genetic algorithms. It also represents a starting point to adding evolvable entities to kinetic market models.
\end{abstract}

%
% Keywords. The author(s) should pick words that accurately describe the work being
% presented. Separate the keywords with commas.
\keywords{optimization heuristic, kinetic market models, econophysics of inequality, diversity, family competition evolutionary algorithm}

%
% This command processes the author and affiliation and title information and builds
% the first part of the formatted document.
\maketitle

\section{Introduction}
Econophysics defines itself as an interdisciplinary discipline that applies methods of physics, especially statistical mechanics, to problems of economics, finance, and other social sciences\cite{schinckus20161996}. One of the main contributions made by this discipline occurs in economic inequality, where econophysicists have been developing kinetic markets \cite{sharma2016physicists}. Although the inspiration originates in the kinetic theory of gases, these market models are constructed using only the concepts of energy exchange and do not have speed or direction. In this metaphor, the concepts of money and individuals replace those of energy and atoms. The atomic collision becomes a monetary transaction, in which a random pair of agents exchange a certain amount of resources guided by a stochastic rule. Using simple rules of exchange, the model evolves this artificial society to identical or similar distributions found in econometric data \cite{chatterjee2004pareto,cerda2013lgem}. Many of these markets reach a stationary distribution, whatever the initial population is. A steady-state arises through similar behavior observed in modern human societies: there is a progressive transfer of almost all resources to a small group or individual. Unlike the phenomenon econophysicists call condensation\cite {hayes2002computing}, when all individuals converge to the same state, the final distribution in those models sustains the population in the interval between none and the sum of all the resources available to society. This dynamic resembles a minimization process because it continually moves the majority of individuals to even smaller states of energy. As noted before \cite {luquini2018rethinking}, the emergence of these distributions and the dynamics of the model suggests an optimization heuristic. Despite the authors' insights, their work failed to provide an efficient algorithm. This work proposes a better solution to understand and use those dynamics recognizing the kinetic market models as an evolutionary algorithm.

\section{Kinetic Market as selection}
\begin{figure}[h!]     
      \includegraphics[width=.40\textwidth]{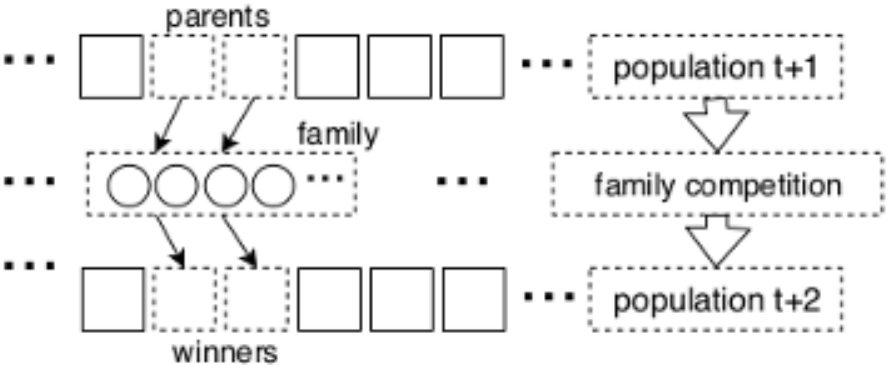}    
     \caption{A schema for family competition or energy exchange.}    
      \label{family}  
\end{figure}

Econophysicists describe their models with an exchange rule,
as equation \ref{eq1} \cite{ispolatov1998wealth,yakovenko2009colloquium}. In addition, there is a further precondition: $(m_i,m_j)$ are pairs of individuals taken at random and without replacement from a population of size $N$ in a generational style.
\begin{subequations} 
  \label{eq1}
       \begin{align}  
        & m_i(t+1) = \epsilon*m_i(t) \hspace{10 mm} \text{ where }  \epsilon=U(0,1) \label{eq2a} \\
        & m_j(t+1) = m_j(t) + ( m_i(t) - m_{i}(t+1)) \label{eq1a}       
      \end{align}
    \end{subequations}
Described that way, the kinetic market model is identical to variations of family competition evolutionary algorithm\cite{mengshoel2008crowding}, as in Fig. \ref{family}. 
The selection for recombination of parents $(m_i,m_j)$ is not biased to the best nor does it have any access to the global state of the population. Moreover, the replacement phase allows the family to challenge only the current parents' position, which is equivalent to the state transition from $( m_i(t),m_j(t) )\rightarrow ( m_i(t+1),m_j(t+1) )$.
 Despite the vocabulary in each discipline, the terms family competition, energy exchange or collision represents the same mechanism: sampling and selection new states starting from $(m_i(t),m_j(t))$. This reasoning allows the populational random search in equation \ref{eq1} to be revealed by its equivalent in \ref{rules1}.
 \begin{subequations}
    \label{rules1}
       \begin{align}  
      & S = \{s_1,s_2,s_3,...s_k\} \text{, $k$ energy states from $U(0,m_i(t))$ } \label{rules1a} \\
      & m_i(t+1) = X, x \in S, P(X=x) = \tfrac{1}{|S|} \label{rules1b}\\
      & m_j(t+1) = m_{j}(t) + (m_{i}(t) -  m_i(t+1)) \label{rules1c} 
   \end{align}
    \end{subequations}
Proposition \ref{rules1a} replaces \ref{eq2a} and defines the sampling set $S$, which has $k$ states with energy less than $m_i(t)$. It implies that $m_i(t+1)$ is a uniformly random choice among possible states of lower energy than $m_i(t)$. The $m_j(t+1)$ is select using the energy conservation principle. In econophysics terms, it means that the loss of agent $ m_i(t)$ in the transaction will be used to determine exactly the energy gain of $m_j(t)$, therefore $m_i(t+1) + m_j(t+1) = m_i(t) + m_j(t)$ is true. \par In conventional application of evolutionary computing, the determination of an individual with energy level, a.k.a fitness, equivalent to the conservation principle is generally not known. For these cases, the non-conservative kinetic model fits better\cite{slanina2004inelastically}. It has the property of scaling down the distribution interval while maintaining its shape. The equation $m_ j(t+1) \approx m_j(t) + (m_ {i}(t) - m_i(t+1))$ replaces \ref{eq1a} and \ref{rules1c}, favoring the interpretation of $m_j(t+1)$ as a result of the sampling process. Consequently, $m_j(t+1)$ becomes an approximation through the selection of a sample closest to the one required for strict energy conservation. These definitions allow a more abstract kinetic market model as described in \ref{rules2}. In this form, the model supports any evolutionary computing individuals.

	\begin{minipage}{.45\textwidth}
  	  \begin{subequations} 
  	  \label{rules2} 
  	  \begin{align} 
  	  & S = \{s_1,s_2,...s_k\}\text{from $f(m_i(t),m_j(t)) \cup \{ m_i(t),m_j(t) \}$  } \label{rules2a} \\
  	   & W = \{  w \in S | w \geq m_j(t) \} \label{rules2d} \\ 
  	   & L = \{  l \in S | l \leq m_i(t) \} \label{rules2e} \\
  	   & m_i(t+1) = X, x \in L, P(X=x) = \tfrac{1}{|L|} \label{rules2b} \\
  	   & Q = \{ q \in W \text{ } | \text{ } q <= m_{j}(t) + (m_{i}(t)-m_i(t+1)) \} \label{rules2e1} \\
  	   & m_j(t+1) = \max Q \label{rules2c} 
  	    \end{align}
  	     \end{subequations}   
  	  \end{minipage}% This must go next to `\end{minipage}`
   \smallskip
   \smallskip
   \par
In evolutionary computing terminology, set $S$ is the family with $k$ members. It has the parents $\{ m_i(t),m_j(t) \}$ and their offsprings created through recombination and mutation by $f(m_i(t),m_j(t))$ .  More precisely, the elements in set $S$ are the family members' fitness. For econophysicists, it is like the asset in their market becomes the solution each agent carries or its intrinsic value.
\section{An exploratory comparison} 
The previous motivated a comparison of the proposed model with other approaches in family competition. For this comparison, the first choice was elitist recombination. It selects the two best individuals with rules \ref{rules5a},\ref{rules5b}. The second option selects the best individual and another random member with rules \ref{rules5a},\ref{rules6a}.
     
  	\begin{minipage}{.20\textwidth}
  	   \begin{subequations}
  	       \label{rules5}
  	      \begin{align}
  	        	    & m_i(t+1)=\min S \label{rules5a}\\
  	        	    & m_j(t+1)=\min B \label{rules5b}\\
  	        	    & \text{for }B={S \setminus m_i(t+1)} 
  	        \end{align}
  	       \end{subequations}
  	  \end{minipage}% This must go next to `\end{minipage}`
  	  \begin{minipage}{.25\textwidth}
  	    \begin{subequations}
  	    \label{rules6}
  	    \begin{align}   
  	    	   & m_j(t+1)=X \label{rules6a}\\   	
  	    	   & \text{where } P(X=x )=\tfrac{1}{|B|} \\
  	    	   & \text{for } x \in B
  	    \end{align} %
  	    \end{subequations}
  	  \end{minipage}
\smallskip \par
\begin{figure}[!h]
  		\includegraphics[width=.40\textwidth]{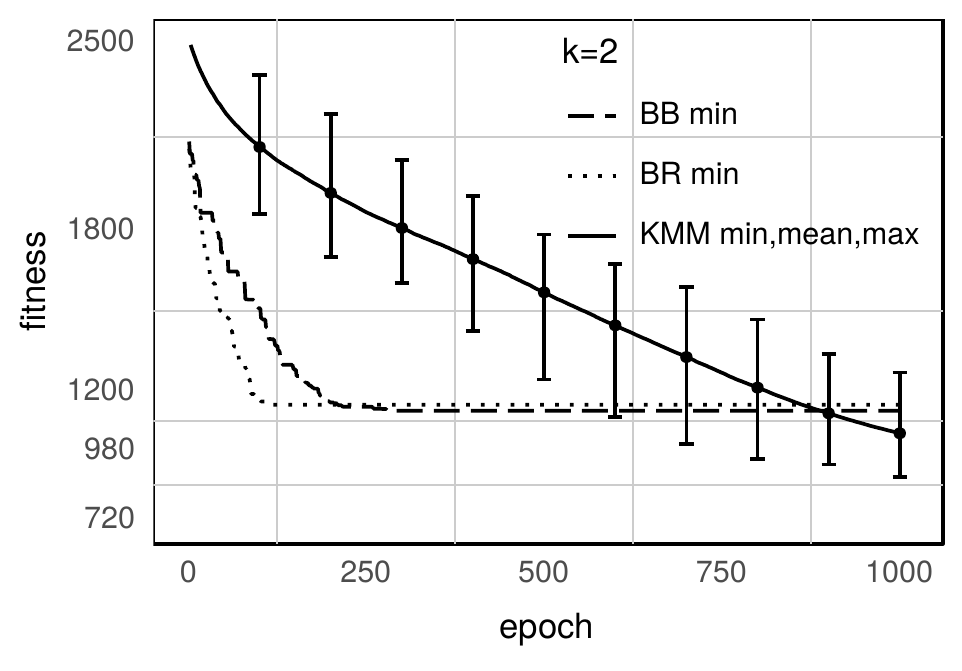}
  		\includegraphics[width=.40\textwidth]{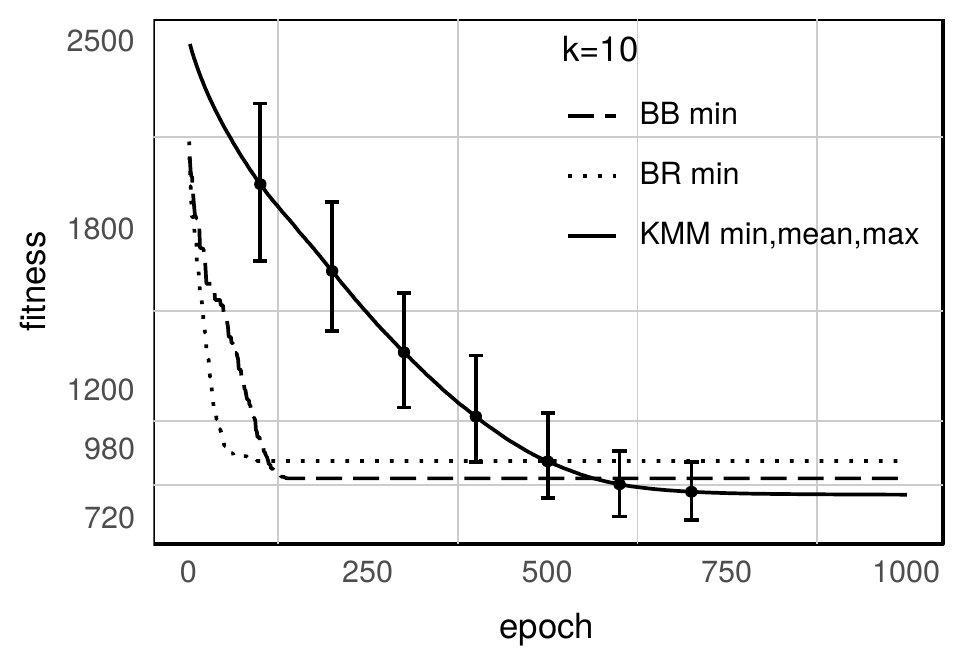}
  		\caption{The optimization of TSPlib's eil76 with the following replacement rules: the two better (BB), the best and other at random (BR) and the kinetic market (KMM).}
  		\label{fig:figure8.1}      			  			
  \end{figure}

  The problem chosen for the comparison was TSPlib's eil76. For each replacement rule, the experiment ran 10 times using a random initial population size of 100. The sampling used just the PMX operator with two configurations: the number of samples k=2 for one experiment and k=10 for another. The results are shown in Fig.\ref{fig:figure8.1}.
 \section{Conclusion} 
 The proposed formulation had a slightly better result in both experiments. The initial data suggest that the kinetic market replacement rule was able to sustain better diversity throughout the simulation compared with the other ones, which lost it very early. In future research, the results should be subjected to a more robust statistical analysis.

\bibliographystyle{acm}
\bibliography{./ref}

\begin{thebibliography}{10}

\bibitem{cerda2013lgem}
{\sc Cerd{\'a}, J., Montoliu, C., and Colom, R.}
\newblock Lgem: A lattice boltzmann economic model for income distribution and
  tax regulation.
\newblock {\em Mathematical and Computer Modelling 57}, 7-8 (2013), 1648--1655.

\bibitem{chatterjee2004pareto}
{\sc Chatterjee, A., Chakrabarti, B.~K., and Manna, S.}
\newblock Pareto law in a kinetic model of market with random saving
  propensity.
\newblock {\em Physica A: Statistical Mechanics and its Applications 335}, 1
  (2004), 155--163.

\bibitem{hayes2002computing}
{\sc Hayes, B.}
\newblock Computing science: Follow the money.
\newblock {\em American Scientist 90}, 5 (2002), 400--405.

\bibitem{ispolatov1998wealth}
{\sc Ispolatov, S., Krapivsky, P.~L., and Redner, S.}
\newblock Wealth distributions in asset exchange models.
\newblock {\em The European Physical Journal B-Condensed Matter and Complex
  Systems 2}, 2 (1998), 267--276.

\bibitem{luquini2018rethinking}
{\sc Luquini, E., and Omar, N.}
\newblock Rethinking exchange market models as optimization algorithms.
\newblock {\em Physica A: Statistical Mechanics and its Applications 491\/}
  (2018), 271--281.

\bibitem{mengshoel2008crowding}
{\sc Mengshoel, O.~J., and Goldberg, D.~E.}
\newblock The crowding approach to niching in genetic algorithms.
\newblock {\em Evolutionary computation 16}, 3 (2008), 315--354.

\bibitem{schinckus20161996}
{\sc Schinckus, C.}
\newblock 1996--2016: Two decades of econophysics: Between methodological
  diversification and conceptual coherence.
\newblock {\em The European Physical Journal Special Topics 225}, 17-18 (2016),
  3299--3311.

\bibitem{sharma2016physicists}
{\sc Sharma, K., and Chakraborti, A.}
\newblock Physicists' approach to studying socio-economic inequalities: Can
  humans be modelled as atoms?
\newblock {\em arXiv preprint arXiv:1606.06051\/} (2016).

\bibitem{slanina2004inelastically}
{\sc Slanina, F.}
\newblock Inelastically scattering particles and wealth distribution in an open
  economy.
\newblock {\em Physical Review E 69}, 4 (2004), 046102.

\bibitem{yakovenko2009colloquium}
{\sc Yakovenko, V.~M., and Rosser~Jr, J.~B.}
\newblock Colloquium: Statistical mechanics of money, wealth, and income.
\newblock {\em Reviews of Modern Physics 81}, 4 (2009), 1703.

\end{thebibliography}

\end{document}